\documentclass[10pt,twocolumn,letterpaper]{article}

\usepackage{cvpr}              

\usepackage[dvipsnames]{xcolor}

\definecolor{cvprblue}{rgb}{0.21,0.49,0.74}
\usepackage{graphicx}
\usepackage{bm}
\usepackage[pagebackref,breaklinks,colorlinks,citecolor=cvprblue]{hyperref}

\def\paperID{7} 
\def\confName{CVPR}
\def\confYear{2024}

\title{Supervised Contrastive Learning for Snapshot Spectral Imaging Face Anti-Spoofing}

\author{\textnormal{Chuanbiao Song}$^{1}$, \textnormal{Yan Hong}$^{1}$, \textnormal{Jun Lan}$^{1}$, \\  \textnormal{Huijia Zhu}$^{1}$, \textnormal{Weiqiang Wang}$^{1}$, \textnormal{Jianfu Zhang}$^{2}$\thanks{Corresponding author.}~\\
$^{1}$Ant Group, 
$^{2}$Qing Yuan Research Institute, Shanghai Jiao Tong University \\
$^{1}${\tt\small songchuanbiao.scb@antgroup.com}, $^{1}${\tt\small yanhong.sjtu@gmail.com}, \\
$^{1}${\tt\small \{yelan.lj, huijia.zhj, weiqiang.wwq\}@antgroup.com}, $^{2}${\tt\small c.sis@sjtu.edu.cn}
}

\begin{document}
\maketitle

\begin{abstract}
This study reveals a cutting-edge re-balanced contrastive learning strategy aimed at strengthening face anti-spoofing capabilities within facial recognition systems, with a focus on countering the challenges posed by printed photos, and highly realistic silicone or latex masks. Leveraging the HySpeFAS dataset, which benefits from Snapshot Spectral Imaging technology to provide hyperspectral images, our approach harmonizes class-level contrastive learning with data resampling and an innovative real-face oriented reweighting technique. This method effectively mitigates dataset imbalances and reduces identity-related biases. Notably, our strategy achieved an unprecedented 0.0000\% Average Classification Error Rate (ACER) on the HySpeFAS dataset, ranking first at the Chalearn Snapshot Spectral Imaging Face Anti-spoofing Challenge on CVPR 2024.
\end{abstract}


\section{Introduction}
\label{sec:intro}
Face recognition technologies~\cite{parkhi2015deep,schroff2015facenet,meng2021magface,kim2022adaface,boutros2022elasticface,bae2023digiface,wu2023face}, with the extensive applications in various aspects of our life like mobile payments and access control systems, has significantly enhanced convenience. 
Nonetheless, its susceptibility to diverse forms of attacks limits its reliable application. Numerous malicious attacks, including the use of printed photos, video replays, and faces with flexible masks, can readily mislead these systems into making wrong judgments. 
To ensure the dependable operation of face recognition systems, face anti-spoofing (FAS) ~\cite{yu2020searching,yang2014learn,yu2020searching,komulainen2013context} methods are crucial for identifying and mitigating various attacks.

Confronted with the challenge of highly convincing silicone or latex masks, the deployment of innovative spectroscopy sensors~\cite{steiner2016reliable,rao2022anti} can notably boost the discriminative power of face recognition systems against these attacks. 
Snapshot Spectral Imaging (SSI)~\cite{cao2016computational,kester2011real,hagen2013review} technologies possess the ability to capture compressed sensing spectral images, positioning it as an effective tool for the integration of spectroscopic information into current face recognition systems. Recently, utilizing a snapshot spectral camera, the Chalearn Snapshot Spectral Imaging Face Anti-spoofing Challenge at CVPR 2024 successfully acquires SSI images of both real and fake faces, and creates the first snapshot spectral face anti-spoofing dataset, named HySpeFAS. 
This dataset encompasses 6760 hyperspectral images, each reconstructed from SSI images using the TwIST~\cite{rueda2019snapshot} algorithm and featuring 30 spectral channels. 
These data present invaluable opportunities for FAS to advance the sophistication and reliability of algorithms.

In this paper, we present our approach tailored for the FAS task. 
We design a re-balanced contrastive learning approach, aimed at capturing the detailed and intrinsic patterns from the imbalanced dataset. 
We embed class-level contrastive learning into FAS task by employing data resampling to mitigate class-level imbalances in the dataset.
Furthermore, we introduce an innovative real face-oriented reweighting methodology to effectively eliminate potential bias to the identity of the face.  
The proposed method achieves 0.0000\% ACER on the HySpeFAS dataset and ranks first place on the Snapshot Spectral Imaging Face Anti-spoofing Challenge at CVPR 2024.

\begin{figure*}[htp]
\begin{center}
\includegraphics[width=1\linewidth]{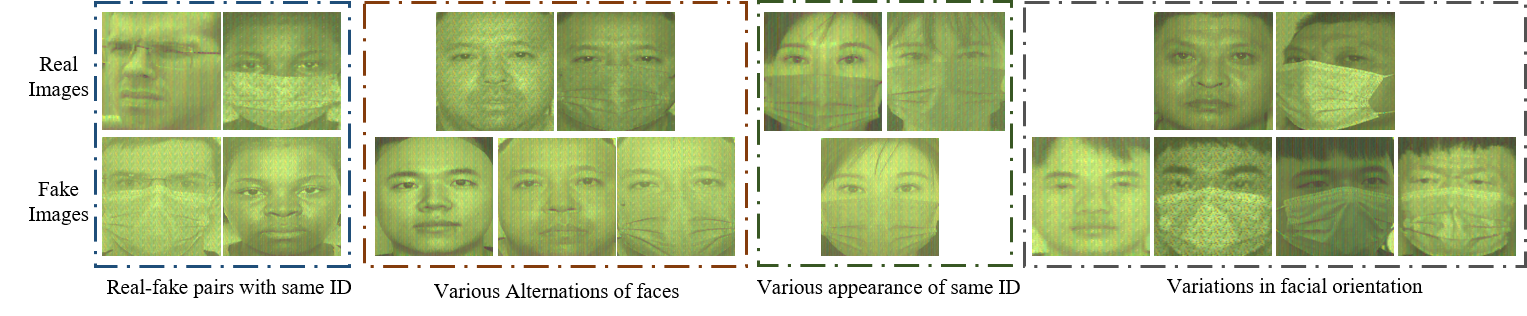}
\end{center}
\caption{Analysis of examples from HySpeFAS dataset. From top to bottom: real images and fake images. From left to right: examples in the first box show the same ID face across real and fake images; the second one indicates the various alternations of real and fake images; the third one visualizes different appearances of images from the same ID; the last one shows different orientations of real images and fake images. }
\label{fig:vis_dataset} 
\end{figure*}

\begin{table}[h]
\centering
\footnotesize
\vspace{4pt}
\setlength{\tabcolsep}{5.8mm}{
\scalebox{0.95}{
\begin{tabular}{l|ccc} 
\toprule
Data & Total & Fake Class & Real Class \\
\midrule
Train  & 3900 & 3380 & 520 \\
Val  & 936 & 728 &  208 \\
Test & 1924 & - & -\\
\bottomrule
\end{tabular}}}
\caption{The split of train/validation/test real images and fake images on the HySpeFAS dataset.}
\label{tab_data_count}
\end{table}

\section{HySpeFAS dataset}
\label{sec:dataset}
The HySpeFAS dataset utilizes a snapshot spectral camera to obtain SSI images of both real and fake faces, and those images are reconstructed from SSI images by TwIST algorithm and characterized with 30 spectral channels.
In total, the dataset provides 6760 hyperspectral images, as detailed in Table~\ref{tab_data_count}. 
For the Snapshot Spectral Imaging Face Anti-spoofing Challenge at CVPR 2024, the organizers have provided all images along with their spectral matrices.
The dataset is divided into a training set with 3900 images and a validation set containing 936 images.
Visual examples from the dataset are shown in Figure~\ref{fig:vis_dataset}, facilitating an in-depth analysis of the HySpeFAS dataset. 
This analysis serves as the foundation for data resampling and reweighting strategies discussed in Section~\ref{sec:method}.

From Table~\ref{tab_data_count}, it is evident that although the dataset is valuable due to the challenging acquisition process, the quantity of data remains limited.
Additionally, the number of counterfeit face images significantly outnumbers that of genuine faces, creating a pronounced imbalance between the two primary classes of images.
In Figure~\ref{fig:vis_dataset}, several key characteristics of the HySpeFAS dataset are identified: 
\textbf{(1).} identical identifiers (ID) are present in both fake and real face images. 
\textbf{(2).} face images exhibit a variety of alterations, including masks and transparent masks. 
\textbf{(3).} faces from the same ID show considerable variation in physical appearance. 
\textbf{(4).} variations in facial orientation and other conditions are also observed.
These characteristics present considerable challenges, and the methodology proposed in this paper is designed to address these specific aspects.
\begin{figure*}[htp]
\begin{center}
\includegraphics[width=1.0\linewidth]{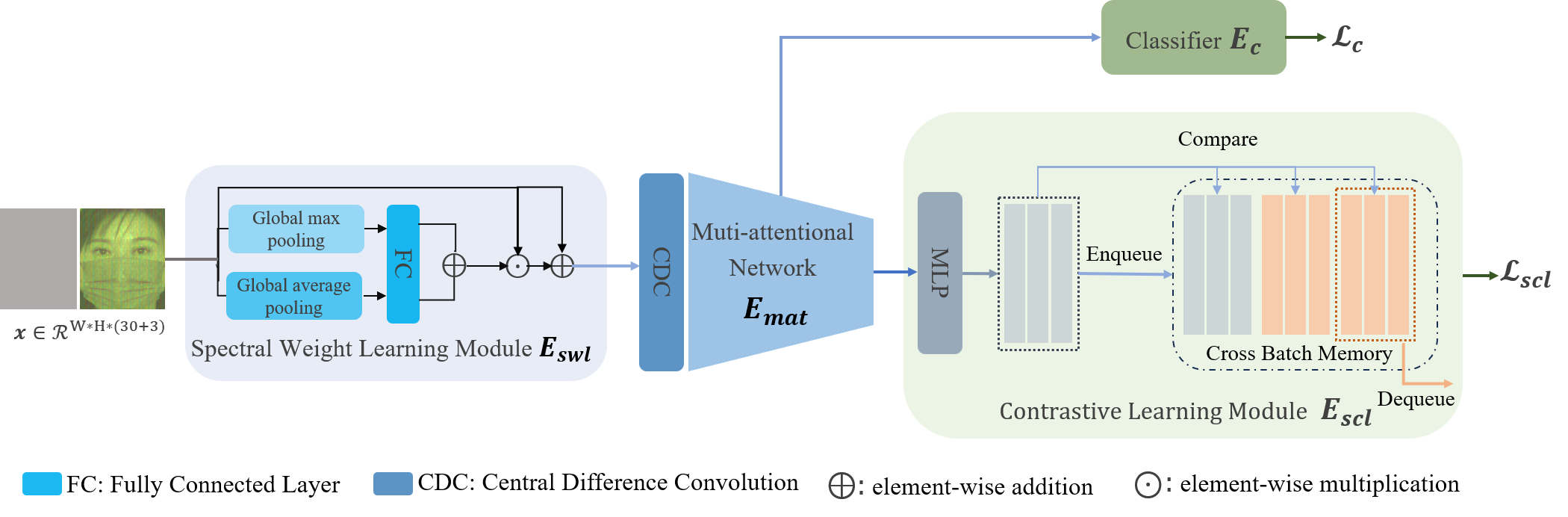}
\end{center}
\caption{The framework of our proposed method.}
\label{fig:framework} 
\end{figure*}

\section{Methodology} \label{sec:method}
In this section, we first present the data preprocessing for the HySpeFAS dataset on the basis of data analysis in Section~\ref{sec:dataset}. 
Then we introduce our framework comprised of multiple modules. 
Next, we describe the used loss function compatible with different modules. 
Finally, the intra-class \textit{mixup}~\cite{ZhangCDL18}, the real face-oriented reweighting, and the cross batch memory~\cite{WangZHS20} are integrated into the training strategy to promote the supervised contrastive learning.

\subsection{Data Preprocessing}
Given the volume and the imbalance between the real and fake sample quantities within the dataset, we initiate our approach with preprocessing enhancements to address this issue.

\paragraph{Class Balancing}
As shown in Table~\ref{tab_data_count}, we analyze the class numbers of the training data and validation data, and can observe that the HySpeFAS dataset is an unbalanced dataset, where the number of fake data is much larger than that of real data. To eliminate the effects of the imbalanced data, we adopt the oversampling strategy to rebalance the data distribution by amplifying the volume of real instances.

\begin{figure}[htp]
\begin{center}
\includegraphics[width=1\linewidth]{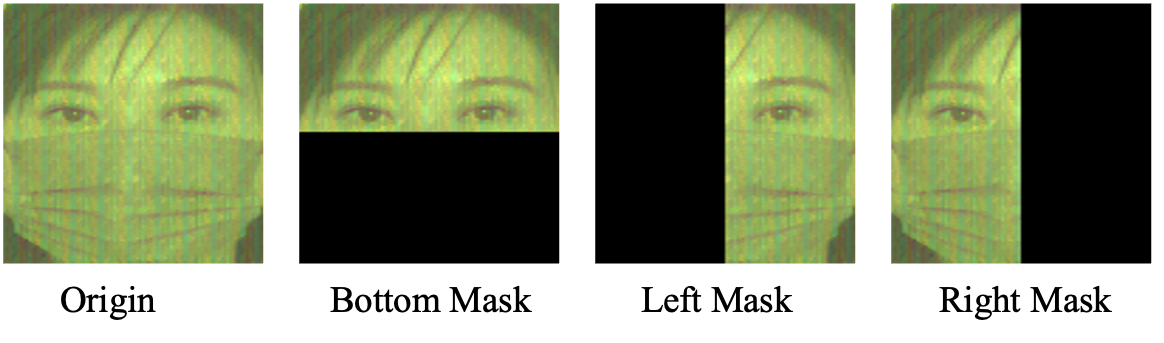}
\end{center}
\caption{Visualization of augmented examples with three different type of \textit{random mask}.}
\label{fig:vis} 
\end{figure}

\paragraph{Data Augmentations}
During the training, we use extensive data augmentations, such as \textit{random crop}, \textit{random horizontal flip}, \textit{cutout}~\cite{abs-1708-04552} and \textit{random mask}. For \textit{random mask}, we randomly mask the bottom half of training samples to eliminate the effect of the worn mask, or randomly mask the left or right half of training samples. Figure~\ref{fig:vis} presents some examples of augmented faces based on \textit{random mask}.

\subsection{Framework}
Our framework, shown in Figure~\ref{fig:framework}, leverage the multi-attention network
MAT~\cite{ZhaoZ0WZY21} as the backbone $\mathbf{E}_{mat}$, and combine the spectral weight learning  module~\cite{FuLZWLZ24} $\mathbf{E}_{swl}$, the central differential convolution~\cite{YuZWQ0LZZ20} $\mathbf{E}_{cdc}$, the classifier $\mathbf{E}_{c}$, and the contrastive learning module $\mathbf{E}_{scl}$. 
image-label pairs from HySpeFAS dataset can be represented as $\{\bm{x}_I,\bm{x}_m,\bm{y}\}$, 
where $\bm{x}_I \in \mathbb{R}^{w\times h\times 3}$  (\emph{resp.,} $\bm{x}_m  \in \mathbb{R}^{w\times h\times 30}$, and $\bm{y} \in \mathbb{R}^{2}$) denotes RGB image (\emph{resp.,} spectral matrix, and one-hot label). We concatenate $\bm{x}_I$ and $\bm{x}_m$ along the third channel to form input sample $\bm{x} \in \mathbb{R}^{w\times h\times 33}$.

Given the multimodal nature of the input, input sample $\bm{x}$ first is processed with the spectral weight learning module $\mathbf{E}_{swl}$ to assign different weights to input channels. 
Then we adopt the $\mathbf{E}_{cdc}$ as the first convolution layer to obtain gradient-level features, and use a multi-attentional network $\mathbf{E}_{mat}$ to learn discriminative features. 
Feded with the learned discriminative features, the classifier $\mathbf{E}_{c}$ is responsible for distinguishing fake faces from real faces, while the contrastive learning module $\mathbf{E}_{scl}$ is dedicated to promoting the discriminability of the learned features.

\subsection{Training Strategies}

\subsubsection{Intra-class \textit{Mixup}} 
Given the limited size of the training dataset, the risk of overfitting is heightened when employing large neural networks. 
To mitigate this and enrich the diversity of training samples, we utilize a variant of \textit{mixup}, inter-class \textit{mixup}. 
As formulated by Eqn.~\ref{eq: mix_up}, inter-class \textit{mixup} generates training samples $(\hat{\bm{x}}, \hat{y})$ by interpolating between two training samples from the same class. 
\begin{equation}
\begin{gathered}
\hat{\bm{x}} = \lambda \cdot {\bm{x}_i} + (1 - \lambda) \cdot \bm{x}_j, \\
\hat{\bm{y}} = \lambda \cdot \bm{y}_i + (1 - \lambda) \cdot \bm{y}_j,\\
\end{gathered}
\label{eq: mix_up}
\end{equation} 
where $\bm{y}_i$ and $\bm{y}_j$ are the one-hot labels of the same class, and $\lambda \in [0, 1]$ represents the mixing parameter. Following the setting of the original \textit{mixup}, we set $\lambda \sim \mathrm {Beta(1.0,1.0)}$.

\subsubsection{Real-face Oriented Reweighting} 
To diminish the model's reliance on content irrelevant to spoofing detection, such as identity and facial features, we introduce Real-face Oriented Reweighting(ROR) strategy during training. 
ROR assigns weights to fake training samples based on their face cosine similarity with real training samples. 
The face cosine similarities are calculated based on the typical face model ArcFace~\cite{DengGXZ19} as follows,
\begin{equation}
\begin{gathered}
w_{\bm{x}^{F}_i} = \max \frac{1 + cos(f_{\mathrm{Arcface}}(\bm{x}^{F}_i), f_{\mathrm{Arcface}}(\bm{x}^{R}_j))}{2},\\
\end{gathered}
\end{equation}
where $\bm{x}^{R}_i$ (\emph{resp.}, $\bm{x}^{F}_j$) denotes the real face image (\emph{resp.},fake face images). 
Fake samples exhibiting higher similarity to real faces are assigned greater weights. 
This reweighting approach encourages the model to deprioritize learning from features strongly tied to identity and facial appearance, thereby focusing more on distinguishing genuine from faces.

\subsubsection{Objective Functions} 

We design two distinct loss functions: one loss for classification, and a contrastive loss for regularizing real and fake data features. 
These losses are combined in a weighted sum to create the overall loss function for training the framework.

\paragraph{Focal Loss.} Instead of using the typical cross entropy for classification, we adopt the focal loss~\cite{LinGGHD17}, which is based on a variant of cross entropy for binary classification:
\begin{equation}
\mathcal{L}_{\text {c}} = \mathrm{FL}(p_t) = -(1-p_t)^{\gamma}log(p_t),
\label{eq: focal_loss}
\end{equation}
where $p_t$ is the probability that the model predicts for the ground truth object and we set $\gamma$ as 2. 
According to Eqn.~\ref{eq: focal_loss}, the focal loss gives less weight to easy examples and gives more weight to hard misclassified examples.

\paragraph{Supervised Contrastive Loss.} The objective of contrastive regularization loss is to optimize the similarity and dissimilarity of real and fake data embeddings. 
The contrastive regulation loss is formulated as:
\begin{equation}
\begin{gathered}
\mathcal{L}_i^{\text {sup }}=\sum_{j=1}^{N} 1_{i \neq j} \cdot 1_{\tilde{y}_i= \tilde{y}_j} \cdot \log \frac{\exp \left(\boldsymbol{z}_i \cdot \boldsymbol{z}_j / \tau\right)}{\sum_{k=1}^{N} 1_{i \neq k} \cdot \exp \left(\boldsymbol{z}_i \cdot \boldsymbol{z}_k / \tau\right)},\\
\mathcal{L}_{\text {scl }}=-\sum_{i=1}^{N} \mathcal{L}_i^{\text {sup }}, 
\end{gathered}
\label{eq: con_loss}
\end{equation}
where $\bm{z}_i$ is denoted as the normalized embedding of the training sample $\bm{x}_i$ from $\mathbf{E}_{scl}$ module, and $\tau$ serves as a temperature hyper-parameter. 

According to Eqn.~\ref{eq: con_loss}, the supervised contrastive loss maximizes the cosine similarity between the training samples with the same category, while simultaneously minimizing the cosine similarity between the training samples with different categories. 
We compute the loss between real and fake samples to encourage the model to learn a generalizable representation that across different images.

To boost the performance of the supervised contrastive loss, we utilize the cross batch memory (XBM)~\cite{WangZHS20} to collect sufficient hard negative pairs for contrastive learning. Specifically, XBM memorizes the embeddings of recent mini-batches and can provide sufficient embeddings for calculating the supervised contrastive loss. It operates on a queue principle, enrolling the latest batch of embeddings while simultaneously removing the oldest, maintaining a dynamic and up-to-date memory bank for optimization.

\textbf{Overall Loss.} The final loss function of the training process is the weighted sum of the above loss functions:
\begin{equation}
\mathcal{L} = \mathcal{L}_{\text {c}} + \lambda_{scl} \cdot \mathcal{L}_{\text {scl}},
\end{equation}
where $\lambda_{scl}$ is a hyper-parameter for balancing the overall loss, and we set $\lambda_{scl}$ as 10 for a strong regularization.

\section{Experiments}
In this section, we will describe the evaluation
metrics, training details, as well as the  performance of our proposed method on the HySpeFAS dataset.

\subsection{Evaluation Metrics}
Following the HySpeFAS dataset, we select the Attack Presentation Classification Error Rate (APCER), Bona
Fide Presentation Classification Error Rate (BPCER), and
Average Classification Error Rate (ACER) as the evaluation metric. Specifically, APCER and BPCER are formulated as below:
\begin{equation}
APCER = \frac{FN}{TP+FN}, \ \ BPCER = \frac{FP}{FP+TN},
\end{equation} 
where FN(False Negative) and FP(False Positive) refer to the number of incorrectly classified fake or real samples respectively, and TP(True Positive) and TN(True Negative) refer to the number of correctly classified real or fake samples respectively.
The ACER is calculated as below:
\begin{equation}
ACER = \frac{APCER+BPCER}{2},
\end{equation}
which is used as the main evaluation metric on the test set
and determine the final ranking of the competition. 
The lower the ACER value, the better the performance.

\subsection{Training Details}
We implement our method on 1 NVIDIA Tesla A100 80G GPU based on open-source framework PyTorch~\cite{paszke2017automatic}. We train the network using the ASAM optimizer~\cite{KwonKPC21} with 30 epochs. The learning rate is 0.01 initially and adjusted by the cosine annealing schedule~\cite{LoshchilovH17}. The batch size is 240 and the weight decay is $5 \times 10 ^{-3}$, and the memory size of the XBM~\cite{WangZHS20} is 1200. The temperature parameter $\tau$  of the supervised contrastive loss is set to 0.07.

\subsection{Performance Results}
We compare the performance of our method and the solutions of other teams on the test set in the Snapshot Spectral Imaging Face Anti-spoofing Challenge at CVPR 2024. 
The evaluation scores of ours and other teams are shown in Table~\ref{tab_score}. We can observe that all the top 10 teams achieve excellent performance results, where all the ACER results are less than 1\%. Our method achieves ACER, APCER, and BPCER by 0\%, 0\% and 0\%, respectively, ranking the first place in this competition.


\begin{table}[h]
\centering
\footnotesize
\vspace{4pt}
\setlength{\tabcolsep}{3.2mm}{
\scalebox{0.95}{
\begin{tabular}{l|ccc} 
\toprule
Team  & ACER(\%) & APCER(\%) & BPCER(\%) \\
\midrule
DXAI & 0.7237 & 0.8065 & 0.6410 \\
ZTT & 0.6927 & 0.7444 &	0.6410 \\
galileo & 0.6359 &	0.3102 &	0.9615 \\
kk\_li & 0.6359 &	0.3102 &	0.9615 \\
ctyun-ai &0.4756  &0.3102&	0.6410   \\
Ricardozzf &0.2223 &0.1241& 0.3205  \\
hexianhua &0.1861 &0.3722 & 0.0000  \\
SeaRecluse &0.0310 &0.0620 & 0.0000  \\
CTEL\_AI & 0.0000 & 0.0000 & 0.0000 \\
\textbf{Ours} & \textbf{0.0000} & \textbf{0.0000} & \textbf{0.0000}  \\ 
\bottomrule
\end{tabular}}}
\caption{The top10 leaderboard of the Snapshot Spectral Imaging Face Anti-spoofing Challenge at CVPR 2024.}
\label{tab_score}
\end{table}



\section{Conclusion}
In this paper, we introduce supervised contrastive learning for snapshot spectral imaging face anti-spoofing based on the multi-attention neural network. Furthermore, to boost the supervised contrastive learning, we utilize the intra-class mixup to improve the diversity of training samples, the real-face oriented sample reweighting to avoid the effects of the identity feature, and the cross-batch memory to increase the number of the contrastive samples. Experimental results show that the proposed method achieves excellent performance and yields the first place among all teams on the recently conducted the Snapshot Spectral Imaging Face Anti-spoofing Challenge at CVPR 2024.

{
    \small
    \bibliographystyle{ieeenat_fullname}
    \bibliography{egbib}
}


\end{document}